\documentclass[letterpaper, 10 pt, conference]{IEEEtran}  

\IEEEoverridecommandlockouts                              

\usepackage{caption}
\usepackage{cite}
\usepackage{amsmath,amssymb,amsfonts}
\usepackage{graphicx}
\usepackage{textcomp}
\usepackage{xcolor}
\usepackage{verbatim}
\usepackage{enumitem}
\usepackage{array}
\usepackage{tablefootnote}
\usepackage{algorithm}
\usepackage{algorithmic}
\usepackage{url}

\newtheorem{theorem}{\textbf{Theorem}}
\newtheorem{lemma}[theorem]{\textbf{Lemma}}

\newtheorem{remark}[theorem]{\textbf{Remark}}

\title{A Direct Semi-Exhaustive Search Method for Robust, Partial-to-Full Point Cloud Registration}

\author{Richard Cheng, Chavdar Papozov, Dan Helmick, Mark Tjersland
\thanks{All authors are with the Toyota Research Institute}
}

\begin{document}

\maketitle
\thispagestyle{empty}
\pagestyle{empty}

\begin{abstract}
Point cloud registration refers to the problem of finding the rigid transformation that aligns two given point clouds, and is crucial for many applications in robotics and computer vision. The main insight of this paper is that we can directly optimize the point cloud registration problem without correspondences by utilizing an algorithmically simple, yet computationally complex, semi-exhaustive search approach that is very well-suited for parallelization on modern GPUs. Our proposed algorithm, Direct Semi-Exhaustive Search (DSES), iterates over potential rotation matrices and efficiently computes the inlier-maximizing translation associated with each rotation. It then computes the optimal rigid transformation based on any desired distance metric by directly computing the error associated with each transformation candidate $\{R, t\}$. By leveraging the parallelism of modern GPUs, DSES outperforms state-of-the-art methods for partial-to-full point cloud registration on the simulated ModelNet40 benchmark and demonstrates high performance and robustness for pose estimation on a real-world robotics problem (\url{https://youtu.be/q0q2-s2KSuA}).
\end{abstract}

\section{Introduction}

Point cloud registration is the problem of computing the rigid transformation that aligns two given point clouds with unknown point correspondences. It is crucial in many robotic applications, including localization and pose estimation. Given two sets of point clouds, the problem is to find the rigid transformation, $\{R, t\} \in SE(3)$, that optimally aligns those point clouds. Because of the widespread importance of this problem, many solutions have been proposed over the past decades.

Most methods break this problem into two stages (often solved iteratively): (1) finding point correspondences, and (2) computing the optimal transformation given the correspondences. By far the most popular approach is Iterative Closest Point (ICP) \cite{Besl1992}. Unfortunately, finding correspondences is extremely difficult without very good initialization, and a few outliers and/or poor correspondences can lead to large errors. Recent learning-based methods have attempted to use data to improve this correspondence matching and/or learn the entire registration pipeline, but this also introduces issues related to generalization. 

The conventional wisdom has been that directly solving the point cloud registration optimization problem (see Problem \eqref{eq:problem}) is intractable. However, with the advent of powerful GPUs, we show that intelligent application of a parallelized, semi-exhaustive search method can be used to solve the point cloud registration optimization problem efficiently on a GPU. Our proposed direct semi-exhaustive search method (DSES) gives us flexibility to minimize over any norm, enabling greater robustness to outliers and partial overlap \cite{Bouaziz2013}. Furthermore, because the method is not data-driven, we get generalization by design.

Note that in this paper, we specifically focus on the partial-to-full point cloud registration setting (i.e. where we are trying to match a partial, observed pointcloud to a full object pointcloud derived from some known model). This is because in the partial-to-full (and full-to-full) setting, point cloud registration can be seen purely as an optimization (see Problem \eqref{eq:problem}), excepting cases of symmetry. However, in the partial-to-partial setting, there are many instances where the optimal solution to this problem provides poor point cloud registration, which necessitates a correspondence-based registration approach (as opposed to a direct optimization approach). Nevertheless, the partial-to-full point cloud registration problem is common in several robotics applications (e.g. object pose estimation), as seen in Section \ref{sec:tool_pose_correction}.


The contributions of this work are as follows:
\begin{itemize}
    \item Introduce a highly parallelizable algorithm, Direct Semi-Exhaustive Search (DSES), for robust, partial-to-full point cloud registration,
    \item Prove the algorithm's optimality in terms of inlier maximization between point clouds, suggesting its effectiveness both theoretically and practically,
    \item Demonstrate high performance and robustness of DSES for partial-to-full point cloud registration leveraging GPUs, outperforming other methods in both simulated and real-world environments.
\end{itemize}

\section{Related Work}
\label{sec:related_work}

\subsection{Classical Registration Methods}

ICP is by far the most popular algorithm for solving rigid registration problems, and involves alternatively (1) finding point cloud correspondences and (2) solving a least-squares problem to compute the alignment \cite{Besl1992}. While this method is computationally efficient, the problem becomes more difficult in the partial-to-full setting or when there are significant outliers (breaking the one-to-one point correspondence assumption). Several ICP variants have been proposed to deal with these issues, for example by introducing point-to-plane correspondences \cite{Yang1992}, setting nearest-neighbor distance thresholds \cite{Rusinkiewicz2001,Bae2008}, introducing different objective functions \cite{rusinkiewicz2019symmetric}, and using probabilistic matching \cite{Segal2010,Koide2021}. This class of solutions is very well studied, and \cite{Rusinkiewicz2001,Pomerleau2015} provide reviews of ICP and its variants. 

While such ICP methods are widely used with many open source implementations, they require significant parameter/threshold tuning and may converge to poor local minima without good initialization. Therefore, several methods have been proposed to tackle global alignment. GO-ICP uses a branch-and-bound method to compute the globally optimal point cloud alignment \cite{Yang2016}. Other methods have been proposed to identify the globally optimal solution through convex relaxation \cite{Maron2016} or mixed-integer programming \cite{Izatt2020}.

However, even globally optimal solutions may yield poor point cloud registration in common real-world settings where there is significant noise and/or only partial alignment. One prominent approach is to use robust functions to reduce the importance of outliers \cite{Trucco1999,Fitzgibbon2003}. More recently, it's been show that this issue can be alleviated by adopting error metrics that promote sparsity of point-wise distance between point clouds (e.g. $L_p$-norm with $p \in (0,1)$) \cite{Bouaziz2013,Yang2020}. However, using these more robust error metrics significantly increase the computational cost, as minimization of the $l_2$-norm enables a closed form solution, which is not the case for $p \in (0,1)$. Other methods have proposed different metrics that can achieve a similar sparsity with faster computation \cite{Zhang2022}.

\subsection{Learning-based Registration Methods}

Recently, several works have looked at incorporating deep learning into the pipeline for point cloud registration. Many of these methods aim to learn/extract point correspondences between point clouds, such that a robust estimator (e.g. RANSAC) can be used to compute alignment \cite{Deng2018a,Deng2018b,Choy2020,Gojcic2019,Kurobe2020}. These correspondences are typically based on keypoint detection with descriptive features \cite{Ao2021,Bai2020,Huang2021}. The challenge lies in the need for repeatable keypoints with highly descriptive features. Therefore, other methods extract correspondences without keypoint detection using ``superpoints'' or hierarchical features \cite{Yu2021,Lee2021,Qin2022}. 

Other works have proposed end-to-end learning to estimate the rigid transformation with a neural network. One class of solutions adopts the same framework as ICP, iteratively establishing soft correspondences and computing the transformation with differentiable weighted SVD \cite{Wang2019a,Wang2019b,Yew2020,Fu2021}. Another class of end-to-end methods aim to extract a global feature vector for each point cloud and regresses the transformation with the global feature vectors \cite{Aoki2019,Huang2020,Zhou2021,Xu2021}. Recently, learning-based graph matching has also been proposed to improve point cloud registration \cite{Saleh2020,Fu2021}.

While these learning-based methods have garnered significant interest, their performance suffers when applied to conditions that are not well-represented in the training set; in such conditions non-learned methods have better generalization ability \cite{Li2021}. In this paper, we suggest that enhanced GPU compute, rather than data-driven learning, may provide an effective and more generalizable avenue to improving point cloud registration.

\section{Problem Setup}
\label{sec:problem}

Consider we are given a source point cloud $\textbf{X} = \{ x_i \in \mathbb{R}^3 ~ | ~ i = 1,...,N \}$ and a reference point cloud $\textbf{Y} = \{ y_j \in \mathbb{R}^3 ~ | ~ j = 1,...,M \}$. Our goal is to compute the rigid transform, $\{R, t\} \in SE(3)$ that optimally aligns the point clouds $\textbf{X}$, $\textbf{Y}$. This can be expressed mathematically as,
\begin{equation}
\underset{R \in SO(3), t \in \mathbb{R}^{3}}{\operatorname{argmin}}  \sum_{x_i \in \mathbf{X}} \min_{y_j \in \textbf{Y}} \| y_j - R x_i - t \|_p
\label{eq:problem}
\end{equation}
where $p$ is any desired norm. While most works consider $p=2$ for computational convenience and speed, researchers have shown that $p\in(0,1]$ or use of a truncated norm provides much more robust results \cite{Bouaziz2013, Yang2020}. Since we are focused on the partial-to-full point cloud registration setting, we consider that $\textbf{Y}$ represents the full object (for which we may have some mesh model) and $\textbf{X}$ may represent only a portion of that object (observed from robot sensors).


\section{Our Method}

We first describe a correspondence-free pure exhaustive search method to solving the point cloud registration problem. While this approach struggles to scale to large problems, it will help us frame/describe our proposed DSES method in Section \ref{sec:DSES}, which shares the same principle but achieves greater efficiency.

\subsection{Pure Exhaustive Search} A pure exhaustive search approach to solving the optimization problem \eqref{eq:problem} would be to discretize over all six rotational/translational DOFs for $\{R, t\} \in SE(3)$, creating a 6D grid of rotations/translations. Given point clouds \textbf{X} and \textbf{Y}, we could then compute the alignment error for each discrete rotation/translation, where the nearest neighbor is computed by iterating over each point-point pair between point clouds \textbf{X} and \textbf{Y}.
\begin{equation}
\textsc{error}(R, t) = \sum_{x_i \in \mathbf{X}} \min_{y_j \in \textbf{Y}} \| y_j - R x_i - t \|_p
\label{eq:error_optimization}
\end{equation}
The $\{R,t\} \in SE(3)$ that yields the minimum error is then our optimal solution. The overall algorithm is extremely simple, and outlined in Algorithm \ref{alg:brute_force}. 

\textbf{The exhaustive search approach outlined in Algorithm \ref{alg:brute_force} is optimal by definition} (up to our discretization resolution) as it iterates through every possible combination in our discrete grid, and allows us to easily optimize over different metrics (not just $L_2$). Therefore, the main reason not to adopt this approach is timing. While the algorithm is highly amenable to GPU parallelization, since we have to discretize over 6 rotational/translational dimensions, and iterate over every point pair in point clouds \textbf{X} and \textbf{Y}, the computation required for this approach scales $O(K^6 M N)$, where $K$ represents the discretization for each dimension, and $N, M$ represent the number of points in point clouds $\textbf{X}, \textbf{Y}$, respectively.

\begin{remark}
    Instead of sampling rotations/translation candidates, one could sample triplets of correspondence pairs and compute the optimal closed form poses for such triplets. However, without decent correspondences this scales poorly (two point clouds with 1000 points leads to $>10$ quadrillion pose candidates). Our correspondence-free approach allows us to brute force our way around correspondences by instead focusing on discretization of 6D pose space.
\end{remark}

\subsection{Direct Semi-Exhaustive Search (DSES)}
\label{sec:DSES}

In this subsection, we boost the efficiency of the pure exhaustive search by iterating only over rotations, $R \in SO(3)$, and efficiently computing the inlier-maximizing translation for each rotation $R$ instead of iterating through every potential rotation/translation combination. We also ensure that this can be easily computed leveraging CUDA kernels. 

Let us define the inlier-maximizing translation in terms of a modified $L_0$-``norm'', $L_0^{m1}$,
\begin{equation}
\begin{split}
    & L_0^{m1}(t | x_i, y_j) = \min ( \| y_j - R x_i - t \|_0 , 1 )  \\
    & \textsc{inliers}(t) = \sum_{x_i \in \textbf{X}} \min_{y_j \in \mathbf{Y}} ~ L_0^{m1} (t | x_i, y_j) .
\end{split}
\label{eq:L0_norm}
\end{equation}
Here $L_0^{m1}$ is simply the $L_0$ ``norm'' saturated at 1. For a given rotation, we can consider the inlier-maximizing translation:
\begin{equation}
t^* = \underset{t \in \mathbb{R}^{3}}{\operatorname{argmin}} \sum_{x_i \in \textbf{X}} \min_{y_j \in \mathbf{Y}} ~ L_0^{m1}(t | x_i, y_j)
\label{eq:translation_optimization}
\end{equation}
As described in Lemma \ref{lemma:translation_optimization} below, the $L_0^{m1}$-optimal (inlier-maximizing) translation $t^*$ for problem \eqref{eq:translation_optimization} can be computed, under mild assumptions, by taking the mode of translations between every \textit{rotated} point-point pair.
\begin{lemma}
Suppose that no points within \textbf{X} are the same, and that no points within \textbf{Y} are the same. Then the optimal solution to \eqref{eq:translation_optimization} is
\begin{equation}
   t^* = \textsc{mode} ( \{ y_j -  R x_i ~ | ~ x_i \in \textbf{X}, y_j \in \textbf{Y} \} ) .
\label{eq:translation_optimization_efficient}
\end{equation}
\label{lemma:translation_optimization}
\end{lemma}

Obviously, the $L_0^{m1}$ norm over continuous points is practically nonsensical. However, since we discretize our points, the $L_0^{m1}$ norm should be considered over a discrete grid. Therefore, we can consider the $L_0^{m1}$-minimizing solution equivalently as the inlier-maximizing solution, given some discretization distance $d$.

\begin{figure}[htbp]
\centerline{\includegraphics[width=0.8\columnwidth]{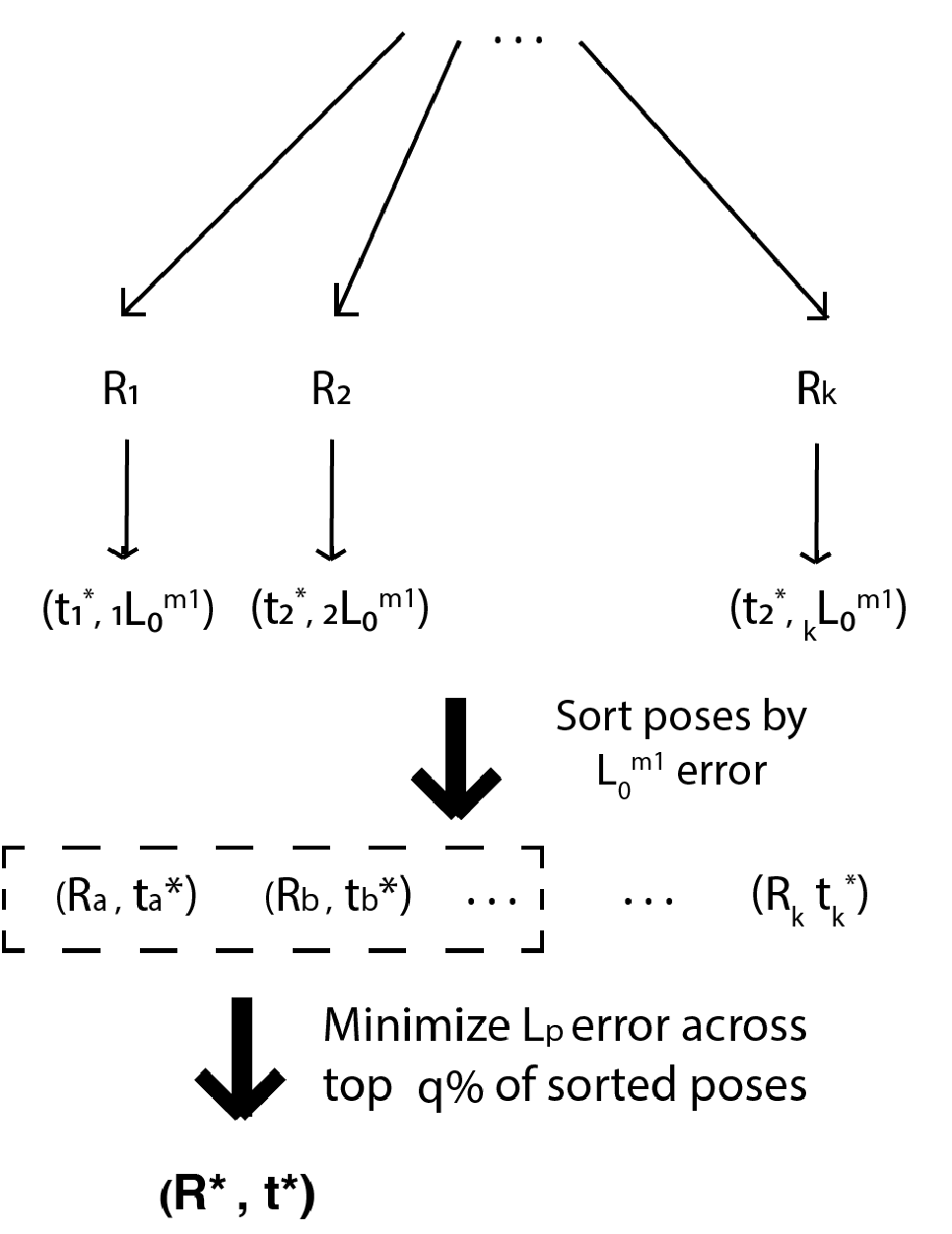}}
\caption{Pictoral description of Algorithm \ref{alg:boo}.}
\label{fig:algorithm}
\end{figure}

Given Lemma \ref{lemma:translation_optimization}, we can iterate over every orientation $R_{\theta, \phi, \xi}$ and use \eqref{eq:translation_optimization_efficient} to compute the corresponding inlier-maximizing translation $t^*_{\theta, \phi, \xi}$ (up to our discretization $\delta_t$). In CUDA, this can be done by counting/storing discrete translation candidates from each point-pair in an array, and taking the argmax of this array. This means we only have to iterate over a 3D grid of orientations, rather than a 6D grid of orientations \textit{and} translations. Once we have our candidate set of rigid transformations $(R_{\theta, \phi, \xi}, t^*_{\theta, \phi, \xi})$, we \textit{sort} them by the $L_0^{m1}$ error; then we compute the optimal rigid transform using our desired $L_p$-norm by brute-force computing the error associated with the best $q\%$ of candidate rigid transforms (as defined by $L_0^{m1}$ error). This process is completely parallelizable and summarized in Algorithm \ref{alg:boo} and illustrated in Figure \ref{fig:algorithm}.

\begin{algorithm}[tb]
	\caption{Pure Exhaustive Search for Registration}\label{alg:brute_force}
	\begin{algorithmic}[1]
	    \STATE \textbf{Input:} $\textbf{X} \in \mathbb{R}^{N \times 3}$, $\textbf{Y} \in \mathbb{R}^{M \times 3}, K \in \mathbb{Z}_{+}, \delta_{t}, \delta_{R} \in \mathbb{R}_{+}$
	    \FOR{$\theta = -K \delta_r; \theta \leq K \delta_r; \theta += \delta_r$}
	    \FOR{$\phi = -K \delta_r; \phi \leq K \delta_r; \phi += \delta_r$}
	    \FOR{$\xi = -K \delta_r; \xi \leq K \delta_r; \xi += \delta_r$}
	    \FOR{$dx = -K \delta_t; dx \leq K \delta_t; \theta += \delta_t$}
	    \FOR{$dy = -K \delta_t; dy \leq K \delta_t; \phi += \delta_t$}
	    \FOR{$dz = -K \delta_t; dz \leq K \delta_t; \xi += \delta_t$}
	    \STATE $e(R_{\theta, \phi, \xi}, t_{dx, dy, dz}) = 0$
	    \FOR {$y_j \in \textbf{Y}$}
	    \STATE $\text{min\_error} = \infty$
	    \FOR {$x_i \in \textbf{X}$}
	    \STATE $\text{error} = \textsc{ERROR} ( y_j - R_{\theta, \phi, \xi} x_i - t_{dx, dy, dz} )$
	    \STATE \text{min\_error} = \textsc{min}(\text{min\_error}, ~\text{error})
		\ENDFOR
		\STATE $e(R_{\theta, \phi, \xi}, t_{dx, dy, dz}) ~~ \text{+=} ~~ \text{min\_error}$
		\ENDFOR
	    \ENDFOR
		\ENDFOR
		\ENDFOR
		\ENDFOR
		\ENDFOR
		\ENDFOR
		\RETURN $(R^{opt}, t^{opt}) = \underset{R, t}{\operatorname{argmin}} ~ e(R, t)$.
	\end{algorithmic}
\end{algorithm}

\begin{theorem}
Suppose that no points within \textbf{X} are within the discretization distance $\delta_t$ of each other, and that no points within \textbf{Y} are within the discretization distance $\delta_t$ of each other. If we choose $L_0^{m1}$ error as our desired $L_p$ norm, then Algorithm \ref{alg:boo} yields the inlier-maximizing solution to problem \eqref{eq:problem} up to our chosen discretization.
\label{theorem:boo}
\end{theorem}

One major benefit of this algorithm is that it is easy to efficiently optimize over any desired metric. Therefore, though Theorem \ref{theorem:boo} suggests the effectiveness of our method, in practice we do not aim solely for inlier-maximization (as this requires fine tuning of our discretization and is sensitive to the noise characteristics of the point clouds). Instead, we use a truncated $L_1$ norm which also has significiant robustness advantages over other norms (see \cite{Yang2020}).

\begin{algorithm}[tb]
	\caption{Direct Semi-Exhaustive Search (DSES)}\label{alg:boo}
	\begin{algorithmic}[1]
	    \STATE \textbf{Input:} $\textbf{X} \in \mathbb{R}^{N \times 3}$, $\textbf{Y} \in \mathbb{R}^{M \times 3}, K \in \mathbb{Z}_{+}, \delta_{t}, \delta_{R} \in \mathbb{R}_{+}, q \in (0,1] $
        \STATE \textcolor{olive}{// For each orientation candidate (parameterized by $\theta, \phi, \xi$), compute the mode $t^*$ (and frequency $M$) of the translation between every point pair up to the specified discretization, $\delta_t$.}
	    \FOR{$\theta = -K \delta_r; \theta \leq K \delta_r; \theta += \delta_r$}
	    \FOR{$\phi = -K \delta_r; \phi \leq K \delta_r; \phi += \delta_r$}
	    \FOR{$\xi = -K \delta_r; \xi \leq K \delta_r; \xi += \delta_r$}
	    \STATE $t^*_{\theta, \phi, \xi} = \textsc{mode} ( \{ \textsc{round} (y_j -  R_{\theta, \phi, \xi} x_i~ , ~ \delta_t) ~ | ~ x_i \in \textbf{X}, y_j \in \textbf{Y} \} )$
        \STATE $M_{\theta, \phi, \xi} = \textsc{count}_{i,j} ( \{ y_j -  R_{\theta, \phi, \xi} x_i + t^*_{\theta, \phi, \xi} < \delta_t \} )$
		\ENDFOR
		\ENDFOR
		\ENDFOR
        \STATE \textcolor{olive}{// Sort the resulting poses $(R_{\theta, \phi, \xi}, t_{\theta, \phi, \xi}^*)$ in decreasing order by frequency $M$, and take only the pose candidates that satisfy $M_{\theta, \phi, \xi} \geq q M^*$.}
        \STATE Sort pose candidates $(R, t^*)$ by count $M$.
        \STATE $M^* = \max_{\theta, \phi, \xi} ~ M$.
        \WHILE{$M_{\theta, \phi, \xi} > q M^*$}
        \STATE \textcolor{olive}{// For each pose candidate, compute the associated error by brute-force checking every nearest neighbor on the GPU.}
        \STATE $e(R_{\theta, \phi, \xi}, t_{dx, dy, dz}) = 0$
        \FOR {$y_j \in \textbf{Y}$}
	    \STATE $\text{min\_error} = \infty$
	    \FOR {$x_i \in \textbf{X}$}
	    \STATE \text{error} =  $\textsc{ERROR} ( y_j - R_{\theta, \phi, \xi} x_i - t^{*}_{\theta, \phi, \xi} )$
	    \STATE $\text{min\_error} = \textsc{min}(\text{min\_error}, ~\text{error})$
	    \ENDFOR
	    \STATE $e(R_{\theta, \phi, \xi}, t^*_{\theta, \phi, \xi}) ~~ \text{+=} ~~ \text{min\_error}$.
	    \ENDFOR
        \ENDWHILE
        \STATE \textcolor{olive}{// Return the pose that minimizes our desired error.}
		\RETURN $(R^{opt}, t^{opt}) = \underset{R, t}{\operatorname{argmin}} ~ e(R, t)$.
	\end{algorithmic}
\end{algorithm}

\section{Results}
\label{sec:results}

All experiments in this section are run with an Intel Core i9-12900K CPU and NVIDIA A6000 GPU. Subsection \ref{sec:results_modelnet} explores the global registration problem on a simulated dataset, and subsection \ref{sec:tool_pose_correction} explores the local registration problem applied to a real-world robotics platform.

\subsection{ModelNet40 Experiments}
\label{sec:results_modelnet}

We conducted experiments on the ModelNet40 benchmark dataset \cite{Wu2015}, which includes 12,311 CAD models from 40 categories, and adapted the experimental conditions from \cite{Yew2020,Fu2021}. As done in these works, we randomly sample 2,048 points from each object, normalized into a unit sphere. To obtain random rigid transformations, we sample three Euler angle rotations in the range $[-45^{\circ}, 45^{\circ}]$ and translations in the range $[-0.5, 0.5]$ on each axis. We transform the source point cloud $\textbf{X}$ using the sampled rigid transform and the task is to register it to the reference point cloud $\textbf{Y}$. 

\textbf{Simulating noise and partial overlap:} For both reference and observation point cloud, we randomly sample 1,024 points \textit{independently} for the source/reference point clouds, and apply a random rigid transformation on the source point cloud. After this, we jitter the points in both point clouds by noise sampled from $\mathcal{N}(0, 0.01)$ and clipped to [-0.05, 0.05] on each axis. Finally, for the source point cloud $\textbf{X}$, we sample a half-space with a random direction $\in \mathcal{S}^2$ and shift it such that $70\%$ of the points are retained, discarding the other $30\%$ to simulate partial-to-full overlap.

~

\noindent
\textbf{Metrics:} We track mean isotropic errors (MIE) of \textbf{R} and \textbf{t} as proposed in \cite{Yew2020}, specified in units of degrees and meters, respectively. We also track mean absolute errors (MAE) of $R$ and $t$ used in DCP \cite{Wang2019a}. Finally, we report the recall with MAE(\textbf{R})$<1^{\circ}$ and MAE(\textbf{t})$<0.1$. The best results are marked in bold font in Tables \ref{table:modelnet_partialfull_unseen} and \ref{table:modelnet_partialfull_rotated}.

We compare our method to state-of-the-art point cloud registration methods RPM-Net \cite{Yew2020}, RGM \cite{Fu2021}, and Predator \cite{Huang2021}. Other methods, such as DCP, PointNetLK, IDAM, DeepGMR, ICP, and FGR \cite{Zhou2016,Aoki2019,Wang2019a, Li2020,Yuan2020} are not directly compared, because experiments in \cite{Yew2020,Fu2021,Huang2021} have already shown that these newer methods have better performance. The ModelNet40 dataset contains official train/test splits for each of the 40 object categories. The first 20 categories were used by the learned methods for training and validation, and the remaining 20 categories were used for testing of all methods. 

As seen from Table \ref{table:modelnet_partialfull_unseen}, DSES outperforms across all metrics, achieving very high recall (i.e. high reliability) and very low errors. The main instances it struggles on are categories with symmetry, where there is ambiguity with respect to some rotational degree-of-freedom. 

A further advantage of DSES is its independence from data (since it is purely solving a minimization problem), making it generalize by design. To highlight this advantage, we repeated the same experiments, but rotated both the reference/source point cloud by the \textit{same} rotation before doing point cloud registration. This effectively modifies the reference frame, but the relative rotation between reference and source point clouds was kept the same. Table \ref{table:modelnet_partialfull_rotated} shows the results of these experiments. We see that the performance of RPM-Net, Predator, and RGM all suffer significantly, while the performance of DSES remains the same. By design, DSES is rotation/translation invariant, and can therefore generalize better to novel conditions.

\begin{table}
\small
\centering
\begin{tabular}{ |p{1.3cm}||p{0.9cm}|p{0.9cm}|p{0.9cm}|p{0.9cm}|p{0.8cm}|}
 \hline
 method & MIE(R) & MIE(t) & MAE(R) & MAE(t) & Recall\\
 \hline
 \hline
 RPM-Net  & 0.98$^\circ$ & 0.011m & 0.51$^\circ$  & 0.005m & 92.5\% \\
 Predator & 1.56$^\circ$ & 0.015m & 0.83$^\circ$ & 0.008m & 82.8\% \\
 RGM      & 1.13$^\circ$ & 0.011m & 0.59$^\circ$ & 0.005m & 92.2\% \\
 DSES     & \textbf{0.72$^\circ$} & \textbf{0.007m} & \textbf{0.36$^\circ$} & \textbf{0.004m} & \textbf{98.1\%} \\
 \hline
\end{tabular}
\caption{Point cloud registration performance on ModelNet40 experiments.}
\label{table:modelnet_partialfull_unseen}
\end{table}

\begin{table}
\small
\centering
\begin{tabular}{ |p{1.3cm}||p{0.9cm}|p{0.9cm}|p{0.9cm}|p{0.9cm}|p{0.8cm}|}
 \hline
 method & MIE(R) & MIE(t) & MAE(R) & MAE(t) & Recall\\
 \hline
 \hline
 RPM-Net  & 1.33$^\circ$ & 0.015m & 0.72$^\circ$  & 0.007m & 88.6\% \\
 Predator & 4.11$^\circ$ & 0.033m & 2.14$^\circ$ & 0.017m & 66.9\% \\
 RGM      & 2.30$^\circ$ & 0.020m & 1.19$^\circ$ & 0.010m & 85.3\% \\
 DSES     & \textbf{0.72$^\circ$} & \textbf{0.008m} & \textbf{0.37$^\circ$} & \textbf{0.004m} & \textbf{97.2\%} \\
 \hline
\end{tabular}
\caption{Point cloud registration performance on ModelNet40 experiments, where reference frame is rotated.}
\label{table:modelnet_partialfull_rotated}
\end{table}

\begin{figure}[htbp]
\centerline{\includegraphics[width=0.95\columnwidth]{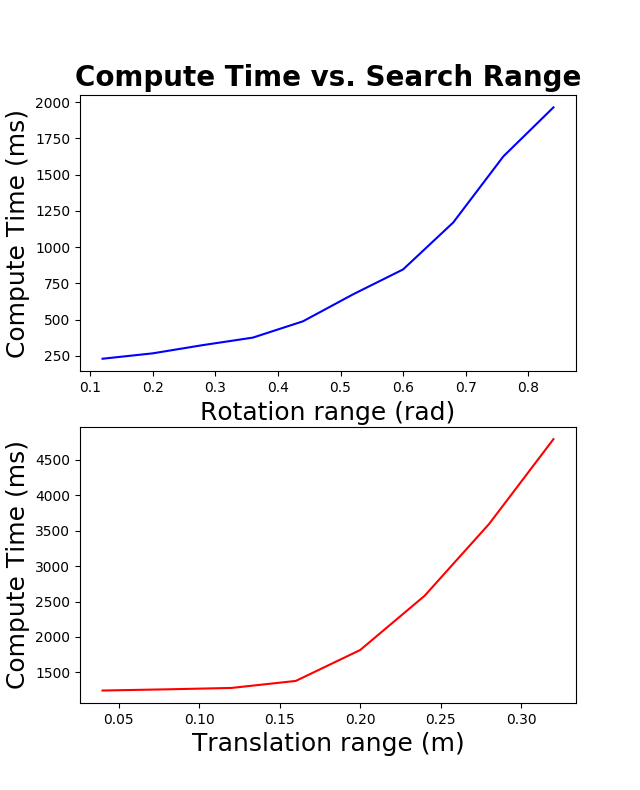}}
\caption{(Top) Plot of computation time versus rotation range, given a translation range of $\pm0.2$m. (Bottom) Plot of computation time versus translation range, given a rotation range of $\pm45^\circ$.}
\label{fig:compute_scaling}
\end{figure}

\noindent
\textbf{Computation Time:} Since our approach is based on a semi-exhaustive search, the computation time is highly dependent on the rotation/translation search range. An analysis of computation time versus search range is shown in Figure \ref{fig:compute_scaling}, and it can be seen that the algorithm is fast when we can constrain the search space by providing an initial guess. Therefore, this method is best for (1) online local point cloud registration with pose initialization, \textit{or} (2) offline global point cloud registration without pose initialization.


\subsection{Real-World Robot Pose Estimation}
\label{sec:tool_pose_correction}

\begin{figure}[htbp]
\centerline{\includegraphics[width=0.98\columnwidth]{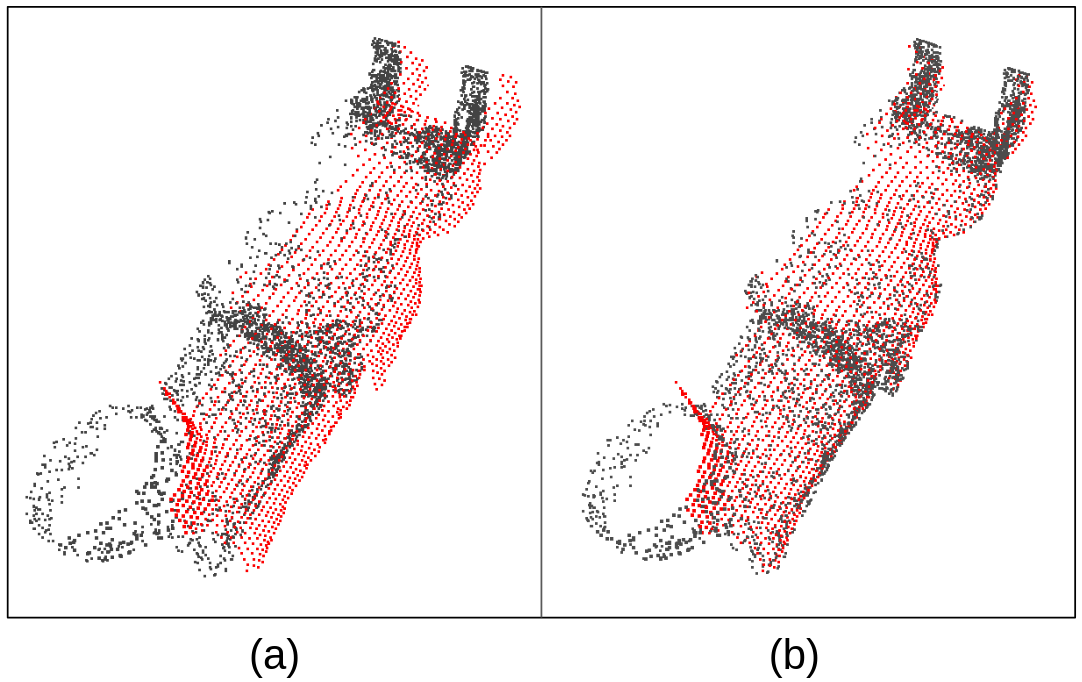}}
\caption{Reference arm tip point cloud (black) vs. observed arm tip point cloud (red) both (a) before pose alignment, (b) after pose alignment with DSES. While the pose delta is small, this difference impacts success of the resulting grasp.}
\label{fig:robot}
\end{figure}

One of our motivations for developing DSES is for robust robot pose estimation. In this subsection, we utilize point cloud registration in order to correct for kinematic errors online in a mobile manipulation robot (see Figure \ref{fig:robot}). The problem is that when we command the robot to a specific joint configuration to achieve a desired gripper pose, error in the kinematics model and robot hardware (e.g. from encoder error, link flexing, etc.) create a discrepancy between the desired and actual gripper pose. To correct for this pose delta, we do point cloud registration between the expected gripper point cloud (obtained from a CAD model) and the perceived gripper point cloud. The perceived gripper point cloud is obtained from a stereo head camera running a learned stereo network to produce dense depth \cite{Shankar2022}. The computed pose delta is used to correct for this error on the robot allowing for more precise gripper positioning  (see supplementary video: \url{https://youtu.be/q0q2-s2KSuA}). This precision is crucial for picking various objects.

\begin{figure}[htbp]
\centerline{\includegraphics[width=0.95\columnwidth]{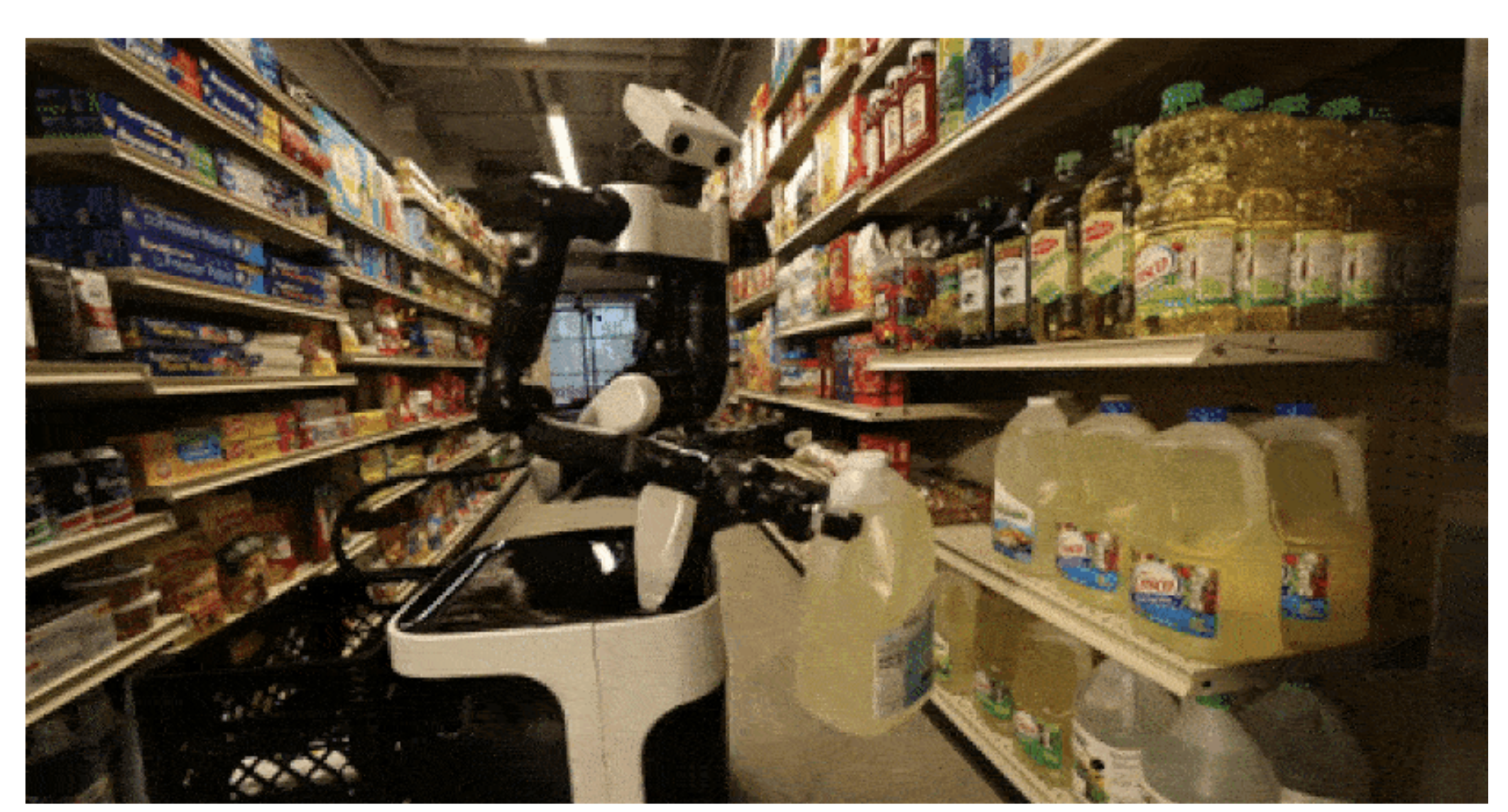}}
\caption{Image of the robot described in Section \ref{sec:tool_pose_correction} picking up a jug from a grocery shelf, after using DSES to correct the gripper pose.}
\label{fig:grocery_robot}
\end{figure}

We ran several experiments of the robot picking items in a real, unmodified grocery store, using point cloud registration running online to correct for kinematic errors in the gripper pose. The search range for DSES was rotations in the range [-5$^{\circ}$, 5$^{\circ}$] and translations in the range [-1.6, 1.6] cm. In these experiments we compare our algorithm, DSES, to the Generalized-ICP algorithm (GICP) \cite{Segal2010} using the C++ implementation from the Point Cloud Library (PCL). As seen from Table \ref{table:robot_results}, our algorithm achieves smaller chamfer distance and much higher success rate, while running significantly faster. Success rate is defined as the percentage of instances where a solution is returned \textit{and} that solution improves (decreases) the chamfer distance. By enabling more precise positioning of the arm tip pose, DSES allows us to successfully grasp items that we otherwise couldn't (e.g. objects with a handle or cap, where high precision is required). The supplementary video shows several instances of such grasps where DSES is crucial for successful grasping.

~

\noindent
\textbf{Note:} These robot experiments test the algorithm under more realistic settings (i.e. more realistic perception noise/artifacts). However, they do not test global registration performance (as the ModelNet40 experiments do), since the problem allows for a decent initial pose based on robot forward kinematics. This initialization also enables the significantly faster computation times.

\begin{table}
\small
\begin{tabular}{ |p{1.2cm}||p{1.6cm}|p{1.6cm}|p{1.5cm}|}
 \hline
 method & Chamfer Distance & Success Rate & Time \\
 \hline
 \hline
 None     & 6.32 mm &  - & - \\
 \hline
 GICP      & 4.69 mm &  68.7\% \newline (167 / 243) & 235 ms \\
 \hline
 DSES     & \textbf{4.36 mm} & \textbf{100\%} \newline (243 / 243) & 50 ms \\
 \hline
\end{tabular}
\caption{Point cloud registration performance for on-robot tool pose correction.}
\label{table:robot_results}
\end{table}

\section{Conclusion}

Historically, point cloud registration has been thought of as an Expectation Maximization problem solved by an iterative process. However, the advent of extremely powerful GPUs allows us to rethink that paradigm. Our proposed DSES algorithm is extremely simple, theoretically optimal, and improves upon state-of-the-art approaches for partial-to-full point cloud registration as shown in Table \ref{table:modelnet_partialfull_unseen}. Our real-world robot experiments in Section \ref{sec:tool_pose_correction} show that DSES deals well with real perception noise/artifacts and is fast for local registration problems, enabling precise robot picking.

The most important advantage of our approach is reliability (i.e. high recall). For most autonomous robotic systems, reliability takes precedence over other metrics - it is better to have an algorithm that gives approximately accurate results 100\% of the time versus perfect results only 90\% and poor results the other 10\%. We believe that the high reliability and generalization of our approach make it ideal for many point cloud registration applications, and as GPUs become cheaper and more powerful (already the Nvidia RTX4090 boasts $>$80 TFLOPS), we believe its advantages will grow.

\subsection*{Current Limitations and Future Work}

While our approach has the advantage of being able to easily trade off between reliability, speed, and problem size, it also means we must be conscious of balancing these three priorities. As seen in Figure \ref{fig:compute_scaling}, our high recall and low errors can come at the cost of higher computation time depending on the pose search range. While increasingly powerful GPUs will alleviate these issues, currently our method is most likely to provide value for (1) offline global registration, or (2) online, local registration. 

As mentioned in Section \ref{sec:problem}, our method optimizes Problem \ref{eq:problem} in order to solve the point cloud registration problem. This works well for partial-to-full (and full-full) point cloud registration, but hits limitations when looking at the partial-to-partial setting. This is because the optimal solution to Problem \ref{eq:problem} can yield poor registration in certain scenarios. We believe incorporating color into the loss function for matching can alleviate this issue for the partial-to-partial setting.

\section{Appendix}

\subsection{Proof of Lemma 2}

For problem \eqref{eq:translation_optimization}, consider the $L_0^{m1}$ defined in \eqref{eq:L0_norm}. By definition, we have:
\begin{equation}
L_0^{m1}(t | x_i, y_j) = 
\begin{cases}
  0 & \text{if $t = y_j - R x_i$}, \\
  1 & \text{otherwise}.
\end{cases}
\end{equation}
This implies that for each $x_i \in \mathbf{X}$, all $M$ translations
\begin{equation}
t_{ij} = y_j - Rx_i, \quad j = 1, \ldots, M
\end{equation}
minimize $L_0^{m1}(t | x_i, y_j)$, where $M$ is the number of points in the point cloud $\mathbf{Y}$. Since $L_0^{m1}(t | x_i, y_j) \in \{0, 1\}$ and no points in $\mathbf{X}$ or $\mathbf{Y}$ are repeated, the following holds
\begin{equation}
\min_{y_j \in \mathbf{Y}} L_0^{m1}(t | x_i, y_j) = 1 - \underset{\forall y_j \in \mathbf{Y}}{\textsc{count}} \big(y_j - R x_i - t \big),
\end{equation}
where we define $\underset{\forall q}{\textsc{count}}(p)$ as the number of times $p = 0$ across all $q$. Therefore, 
\begin{equation}
\sum_{x_i \in \mathbf{X}} \min_{y_j \in \mathbf{Y}} ~ L_0^{m1}(t | x_i, y_j) = N - \underset{\forall x_i \in \mathbf{X}, \forall y_j \in \mathbf{Y}}{\textsc{count}} \big(y_j - R x_i - t \big) .
\end{equation}
where $N$ is the number of points in $\mathbf{X}$. This implies that the following optimization problems are equivalent,
\begin{equation}
\begin{split}
\underset{t \in \mathbb{R}^{3}}{\operatorname{argmin}} \sum_{x_i \in \mathbf{X}} &\min_{y_j \in \mathbf{Y}} ~ L_0^{m1}(t | x_i, y_j) \\
&= \underset{t \in \mathbb{R}^{3}}{\operatorname{argmax}} \underset{\forall x_i \in \mathbf{X}, \forall y_j \in \mathbf{Y}}{\textsc{count}} \big(y_j - R x_i - t \big). \\
 &= \textsc{mode} ( \{ y_j -  R x_i ~ | ~ x_i \in \textbf{Y}, y_j \in \textbf{Y} \} ).
\end{split}
\end{equation}
where the last equality holds by definition of the mode. Therefore, we can conclude that the solution \eqref{eq:translation_optimization_efficient} optimizes problem \eqref{eq:translation_optimization}.

\subsection{Proof of Theorem 3}

As seen from Lines 6-7 of Algorithm \ref{alg:boo}, for every rotation, we compute the accompanying $L_0^{m1}$ optimal translation based on the \textsc{mode} \eqref{eq:translation_optimization_efficient}. Lemma \ref{lemma:translation_optimization} proves that this translation $t^*_{\theta, \phi, \xi}$ is inlier-maximizing for a given rotation, $R$. In addition to computing the translation, $t^*_{\theta, \phi, \xi} = \textsc{mode} ( \{ y_j -  R x_i ~ | ~ x_i \in \textbf{Y}, y_j \in \textbf{Y} \} )$ , we also compute  $M_{\theta, \phi, \xi}$. From the proof of Lemma \ref{lemma:translation_optimization}, we know that the $L_0^{m1}$-norm can be computed as
\begin{equation}
\begin{split}
\sum_{x_i \in \mathbf{X}} L_0^{m1}(t, x_i) &= N - \underset{\forall x_i \in \mathbf{X}, \forall y_j \in \mathbf{Y}}{\textsc{count}} \big(y_j - R x_i - t^*_{\theta,\phi,\xi} \big) \\
&= N - M_{\theta, \phi, \xi} .
\end{split}
\end{equation}
Therefore, the inlier-maximizing ($L_0^{m1}$-minimizing) solution is the one that maximizes $M_{\theta, \phi, \xi}$. In Line 12-13 of Algorithm \ref{alg:boo}, we sort through all pose candidates $(R, t^*)$ to obtain the pose $(R^{opt}, t^{opt})$ that maximizes the count $M_{\theta, \phi, \xi}$, thereby maximizing the number of inliers.

Note that Lines 14-25 are unnecessary for computing the inlier-maximizing solution, but as discussed below, in practice we may desire optimization over a different metric. The post-search steps in Lines 14-25 are necessary to optimize over different norms.

\bibliographystyle{IEEEtran}
\bibliography{references}

\end{document}